\documentclass[final,3p,12pt]{elsarticle}

\usepackage{amssymb}

\usepackage[flushleft]{threeparttable}
\usepackage{adjustbox}
\usepackage{amsmath}
\usepackage{array}
\usepackage{bbm}
\usepackage{bm}
\usepackage{geometry}
\usepackage{hyperref}
\usepackage{multirow}
\usepackage{url}
\usepackage{enumitem}

\newenvironment{myitemize}
{ \begin{itemize}
    \setlength{\itemsep}{0.25\baselineskip}
    \setlength{\parskip}{0pt}
    \setlength{\parsep}{0pt}     
    \vspace*{-0\baselineskip}}
{   \vspace*{-0.5\baselineskip}
  \end{itemize}                  } 

\makeatletter
\newcommand*{\centerfloat}{%
  \parindent \z@
  \leftskip \z@ \@plus 1fil \@minus \textwidth
  \rightskip\leftskip
  \parfillskip \z@skip}
\makeatother

\journal{Current Opinion in Structural Biology}

\begin{document}

\begin{frontmatter}


  \title{Deep Generative Modeling for Protein Design}


  \author[label1]{Alexey Strokach}
  \author[label1,label2,label3]{Philip M. Kim\corref{cor1}}
  \ead{pi@kimlab.org}

  \affiliation[label1]{
    organization={Department of Computer Science, University of Toronto},
    addressline={40 St. George Street},
    city={Toronto},
    postcode={M5S 2E4},
    state={Ontario},
    country={Canada}}

  \affiliation[label2]{
    organization={Donnelly Centre for Cellular and Biomolecular Research, University of Toronto},
    addressline={160 College Street},
    city={Toronto},
    postcode={M5S 3E1},
    state={Ontario},
    country={Canada}}

  \affiliation[label3]{
    organization={Department of Molecular Genetics, University of Toronto},
    addressline={1 King's College Circle},
    city={Toronto},
    postcode={M5S 1A8},
    state={Ontario},
    country={Canada}}

  \cortext[cor1]{Corresponding author.}

  \begin{abstract}
    Deep learning approaches have produced substantial breakthroughs in fields such as image classification and natural language processing and are making rapid inroads in the area of protein design. Many generative models of proteins have been developed that encompass all known protein sequences, model specific protein families, or extrapolate the dynamics of individual proteins. Those generative models can learn protein representations that are often more informative of protein structure and function than hand-engineered features. Furthermore, they can be used to quickly propose millions of novel proteins that resemble the native counterparts in terms of expression level, stability, or other attributes. The protein design process can further be guided by discriminative oracles to select candidates with the highest probability of having the desired properties. In this review, we discuss five classes of generative models that have been most successful at modeling proteins and provide a framework for model guided protein design.
  \end{abstract}

  \begin{graphicalabstract}
  \end{graphicalabstract}

  \begin{highlights}
    \item Machine learning is becoming a key component of the protein design process
    \item Deep generative models can produce novel protein sequences and structures
    \item Conditioned generative models can produce proteins with specific properties
    \item Discriminative oracles can be used to further fine-tune the design process
  \end{highlights}

\end{frontmatter}



\section{Introduction}
\label{sec:introduction}

\begin{figure}[tbp]
  \vspace*{-1in}
  \centering
  \includegraphics[width=1.0\textwidth]{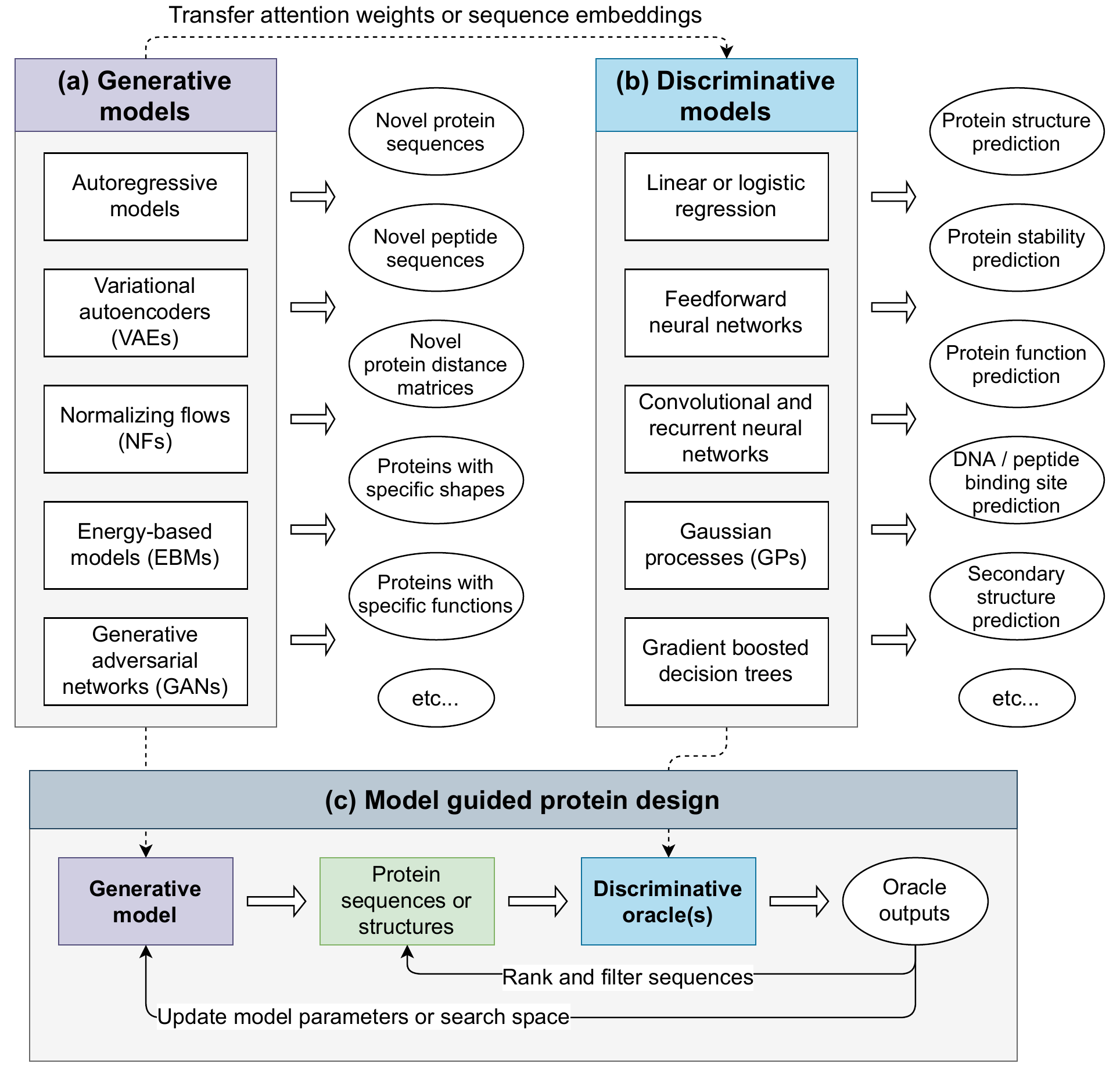}
  \caption{Overview of different machine learning approaches for protein design. \textbf{(a)} Generative models trained on protein sequences or structures can learn probability density landscapes defined by the training data and can be used to generate new proteins that are expressed, are stable and, optionally, have a specific structure and function. These models can loosely be subdivided into self-supervised models, which use supervised learning to learn one aspect of the data from another, and latent variable models, which learn a mapping from latent variables with defined structure to the data. \textbf{(b)} Discriminative models learn to predict specific properties of a protein from its sequence or structure. In some cases, especially when limited training data are available, discriminative models can show a substantial boost in performance when they leverage protein representations learned by the generative models. \textbf{(c)} Model guided protein design involves a generative model, which proposes new protein sequences or structures, and one or more discriminative oracles, which assign a score to the proposed proteins based on their predicted ability to meet specific objectives.}
  \label{fig:overview}
\end{figure}

The optimization of existing proteins, the generation of new proteins with specific shapes and functions, as well as other aspects of protein design, remain key challenges in structural biology. Traditionally, computational protein design has been carried out using tools which use different sampling techniques to explore the energy landscape defined by molecular mechanics force fields or semi-empirical energy functions \cite{huangComingAgeNovo2016}. Recently, the use of machine learning and artificial intelligence has led to breakthroughs in a number of areas \cite{gaoDeepLearningProtein2020, wuProteinSequenceDesign2021, alquraishiMachineLearningProtein2021}, including the accurate prediction of protein structures using AlphaFold2 \cite{jumperHighlyAccurateProtein2021, alquraishiMachineLearningProtein2021}. Given the continued growth in the number of available protein sequences and structural information, machine learning approaches are also posed to become indispensable for efficient and successful protein design.

The goal of this review is to formalize the different machine learning algorithms used for protein design and to delineate a framework for generating novel proteins meeting specific objectives (\autoref{fig:overview}). We first describe five classes of generative models that can be used to produce new protein sequences and structures or learn meaningful representations thereof (\autoref{fig:overview}a). Where appropriate, we also describe supervised models that achieved superior performance because they leveraged representations learned by the trained generative models (\autoref{fig:overview}b). Supervised models, however, are not the focus of this review, as they have been described in detail elsewhere \cite{gaoDeepLearningProtein2020} and their utility is well-established. Finally, we describe strategies that have been used to combine generative models with supervised models, simulations, and domain expertise, to produce the desired proteins (\autoref{fig:overview}c).

\section{Generative models of protein sequences and structures}
\label{sec:self-supervised-models}

\begin{figure}[tbp]
  \centering
  \includegraphics[width=1.0\textwidth]{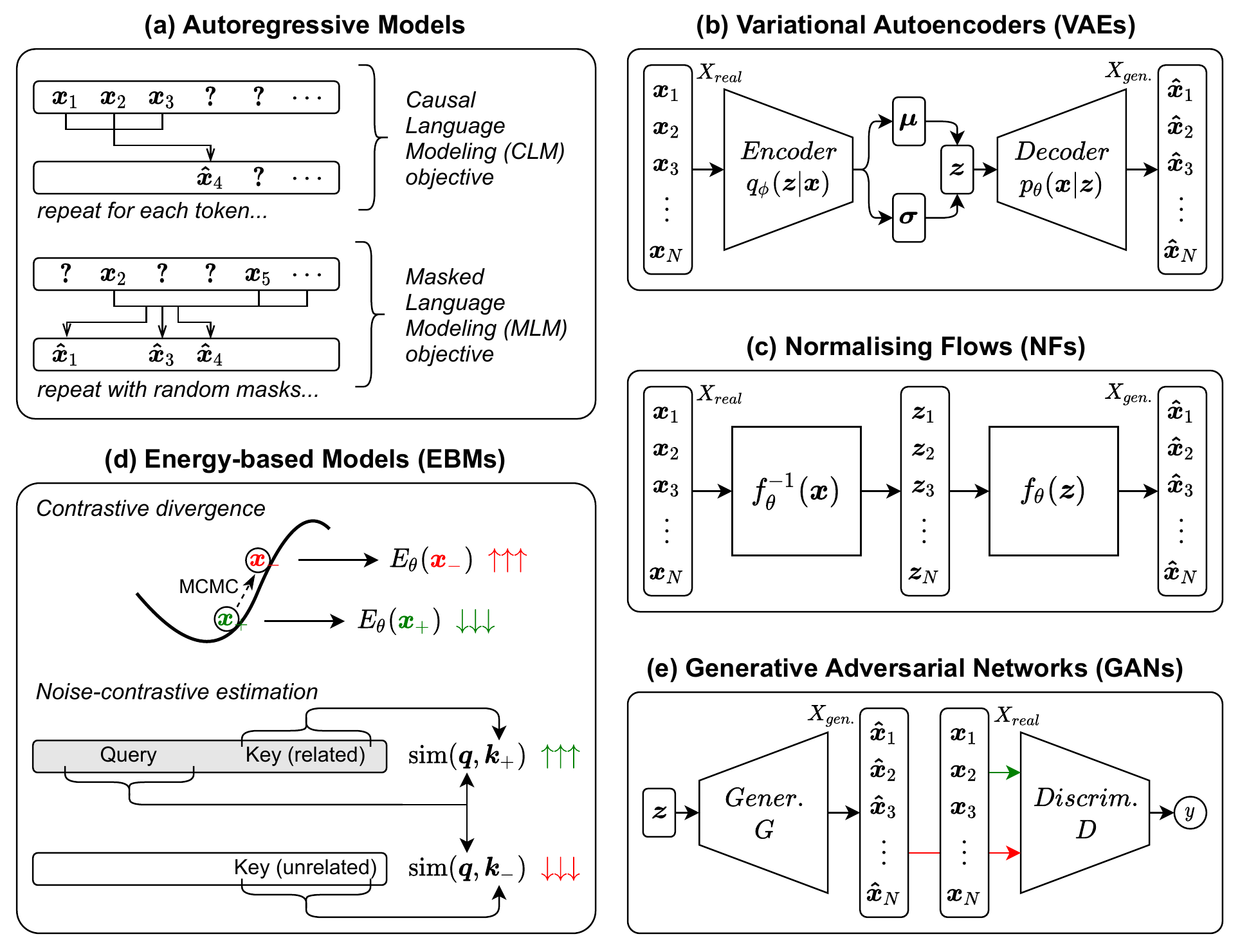}
  \caption{Overview of the five generative model architectures that are covered in this review. \textbf{(a)} Autoregressive models learn to predict the identities of tokens (e.g. amino acids) making up a protein from the identities of preceding or surrounding tokens. \textbf{(b)} Variational autoencoders (VAEs) comprise an encoder trained to parameterize the distribution over the latent variables $\boldsymbol{z}$ and a decoder trained to reconstruct the inputs using samples from the distribution defined by the encoder. \textbf{(c)} Normalizing flows (NFs) use a bijective model to map inputs to and from a latent representation. The model parameters are optimized such that the probability of the training data in the latent space is high while the amount of ``warp'' required to map the data back to the input space is low. \textbf{(d)} Energy-based models (EBMs) learn an energy function that assigns low energies to probable states, including the training data, and high energies to improbable states, often generated by perturbing the training data. Noise-contrastive estimation (NCE) is a training strategy for EBMs where fake examples are sampled from a predefined distribution, and the model is trained to distinguish between real and fake examples. \textbf{(e)} Generative adversarial networks (GANs) comprise a generator trained to produce examples which appear real to the discriminator and a discriminator trained to distinguish between real and generated examples.}
  \label{fig:generative-models}
\end{figure}

\begin{table}[tbp]
  \vspace*{-1in}
  \centering
  \centerfloat
  \begin{threeparttable}
    \caption{Advantages and disadvantages of different generative models used for protein design.}
    \label{tab:model-comparison}

    \footnotesize
    \setlength{\leftmargini}{0.4cm}
    \begin{tabular}{| b{1.9em} |  m{0.5\linewidth} | m{0.5\linewidth} | }
      \hline

      \rule[-0.7\baselineskip]{0pt}{2\baselineskip}  & \multicolumn{1}{c|}{\textbf{Advantages}} & \multicolumn{1}{c|}{\textbf{Disadvantages}} \\
      \hline

      \rotatebox[origin=c]{90}{\textbf{Autoregressive}}
      \rotatebox[origin=c]{90}{\textbf{models}}
      &
      \begin{myitemize}
        \item Native support for modeling categorical variables*.
        \item Native support for modeling data of varying dimensionality.
        \item Gives the exact log-likelihood of the data.
        \item Training is relatively stable.
        \item Can be applied to graph-structured data.
      \end{myitemize} &
      \begin{myitemize}
        \item Limited support for modeling continuous variables* (requires discretization, etc.).
        \item No support for domains that cannot be modeled as sequences or graphs.
        \item Inference is performed one token at a time, making it slow.
      \end{myitemize} \\
      \hline

      \rotatebox[origin=c]{90}{\raisebox{-1\normalbaselineskip}[0pt][0pt]{\textbf{VAEs}}}
      &
      \begin{myitemize}
        \item Native support for modeling categorical and continuous variables*.
        \item Gives the lower bound (ELBO) of the log-likelihood of the data.
      \end{myitemize} &
      \begin{myitemize}
        \item Generated examples tend to be more blurry than with GANs.
        \item Limited support for modeling data of varying dimensionality.
      \end{myitemize} \\
      \hline

      \rotatebox[origin=c]{90}{\textbf{Normalizing}}
      \rotatebox[origin=c]{90}{\textbf{flows}}
      &
      \begin{myitemize}
        \item Native support for modeling continuous variables*.
        \item Gives the exact log-likelihood of the data.
      \end{myitemize} &
      \begin{myitemize}
        \item Little to no support for modeling categorical variables*.
        \item Deeper models, with more parameters, are needed to achieve performance comparable to VAEs / GANs.
        \item The underlying model needs to be bijective.
        \item Calculating the trace of the Jacobian matrix can be computationally expensive.
      \end{myitemize} \\
      \hline

      \rotatebox[origin=c]{90}{\textbf{Energy-based}}
      \rotatebox[origin=c]{90}{\textbf{models}}
      &
      \begin{myitemize}
        \item Native support for modeling categorical and continuous variables*.
        \item Flexible in terms of the types of data that can be modeled.
      \end{myitemize} &
      \begin{myitemize}
        \item Obtaining negative examples can be difficult and computationally expensive.
        \item The model can be biased by the sampling strategy used to generate negative examples.
        \item Calculating probabilities requires evaluating every possible alternative, which can be exceedingly slow or intractable.
        \item Generation is performed through sampling (e.g. MCMC).
      \end{myitemize} \\
      \hline

      \rotatebox[origin=c]{90}{\raisebox{-1\normalbaselineskip}[0pt][0pt]{\textbf{GANs}}}
      &
      \begin{myitemize}
        \item Native support for modeling continuous variables*.
        \item Can generate the most realistic examples (at least in the case of images).
      \end{myitemize} &
      \begin{myitemize}
        \item Special care is needed to model categorical variables* (e.g. using the straight-through Gumbel max trick \cite{maddisonConcreteDistributionContinuous2017,jangCategoricalReparameterizationGumbelSoftmax2017} to back-propagate through outputs).
        \item Training can be less stable than with other models.
        \item Generated examples can be of low diversity (mode collapse).
        \item No native support for mapping existing data to latent space or calculating log-likelihoods.
        \item Limited support for modeling data of varying dimensionality.
      \end{myitemize} \\
      \hline

    \end{tabular}

    \begin{tablenotes}
      \footnotesize
      \item *In the context of protein design, DNA and amino acid sequences are usually represented as \textit{categorical} variables, while protein structures are usually represented as \textit{continuous} variables. This is an important factor that needs to be considered when evaluating models for a particular application.
    \end{tablenotes}

  \end{threeparttable}
\end{table}

Deep generative models have gained wide adoption in recent years due to their ability to learn from massive unlabeled datasets, produce meaningful representations and density estimates of the data and generate new examples of striking coherence. Many generative models have been developed \cite{bond-taylorDeepGenerativeModelling2021} (\autoref{fig:generative-models}), with various trade-offs and limitations that make them well-suited to different aspects of protein design (\autoref{tab:model-comparison}). Here we provide an overview of five classes of deep generative models, including their notable applications to protein design. A more detailed description of each model class can be found in \ref{sec:model-descriptions}.

\subsection{Autoregressive models}

Autoregressive models are trained to predict the next token given the previous tokens or to predict masked tokens given unmasked tokens (\autoref{fig:generative-models}a). These models have received much attention in recent years due to their success in natural language processing (NLP). Similar to language, protein sequences are readily represented as a succession of tokens, which has allowed state-of-the-art NLP models to be applied to protein sequences with few modifications.


The first autoregressive models to be applied to protein sequences used recurrent neural networks (RNNs) with long short-term memory (LSTM) layers \cite{alleyUnifiedRationalProtein2019a, biswasLowProteinEngineering2021} or dilated convolutions \cite{shinProteinDesignVariant2021} to predict the identity of an amino acid given the preceding amino acids. More recently, the transformer architecture \cite{vaswaniAttentionAllYou2017} has gained popularity \cite{raoEvaluatingProteinTransfer2019,elnaggarProtTransCrackingLanguage2020,rivesBiologicalStructureFunction2021,raoMSATransformer2021}, as it generally produces higher reconstruction accuracies and better performance on downstream tasks, including remote homology detection, secondary structure prediction, contact prediction \cite{rivesBiologicalStructureFunction2021, bhattacharyaSingleLayersAttention2020a, raoTransformerProteinLanguage2020}, and mutation effect prediction \cite{meierLanguageModelsEnable2021}. The power of transformers comes primarily from their use of multi-head attention, which allows every element to have direct access to the information stored in every other element in the sequence. Furthermore, transformer models are more efficient to train than RNNs, as they process entire sequences in parallel rather than one element at a time.


Graph neural networks can be viewed as an extension to the transformer architecture \cite{fuchsSETransformers3D2020, bronsteinGeometricDeepLearning2021}, allowing the use of ``edge'' attributes encoding the relationships between pairs of tokens, and making it possible to limit the neighbors with which the information is shared in multi-head attention. \citet{ingrahamGenerativeModelsGraphBased2019a} developed Structured Transformer, a graph neural network with an encoder-decoder architecture, where the encoder takes as input the protein structure, defined by the backbone torsion angles and the distances and relative translations and rotations between pairs of residues, and the decoder generates the amino acid sequences with self-attention to the preceding residues and attention to the embeddings generated by the encoder. Structured Transformer assigns native amino acids higher probabilities than sequence-only autoregressive models, and it is able to recover correct amino acids in NMR protein structures with higher accuracy than Rosetta \cite{lemanMacromolecularModelingDesign2020}. \citet{strokachFastFlexibleProtein2020} developed ProteinSolver, a graph neural network where the input node and edge attributes define the identities and the distances between pairs of amino acids, respectively, and the network is trained to reconstruct the identities of masked amino acids. ProteinSolver generates novel sequences that fold into stable proteins with the desired topologies, as confirmed by an array of computational validation techniques and the circular dichroism spectra of expressed and purified proteins, and it is better able to predict changes in protein stability and affinity than transformers that do not leverage structural information \cite{strokachELASPIC2EL2Combining2021}.

The transformer architecture has also been extended in other ways. \citet{raoMSATransformer2021} used axial attention to leverage the information present in multiple sequence alignments, improving the reconstruction accuracies and achieving better performance on contact prediction, secondary structure prediction, and mutation effect prediction \cite{raoMSATransformer2021, meierLanguageModelsEnable2021}. \citet{madaniProGenLanguageModeling2020, madaniDeepNeuralLanguage2021} used an additional input token to encode the function of the protein and trained a conditional transformer to generate novel protein sequences with predetermined functions. The authors validated their model by generating novel lysozymes and showed experimentally that the generated proteins have lysozyme activity and fold into structures that are characteristic of existing lysozymes. Finally, fine-tuning a pre-trained model using sequences of proteins with the target function or topology can be a simple way of improving model performance on tasks that are specific to those proteins \cite{alleyUnifiedRationalProtein2019a, meierLanguageModelsEnable2021}.

\subsection{Variational autoencoders (VAEs)}

Variational autoencoders (VAEs) use an encoder network to map the inputs to a low-dimensional latent space and a decoder network to reconstruct the inputs using a sample from that latent space (\autoref{fig:generative-models}b). VAEs are trained to minimize the distance between the original and the reconstructed inputs while constraining the latent space to approximate a standard Gaussian to improve generalizability.


VAEs were some of the first unsupervised methods for mutation effect prediction \cite{riesselmanDeepGenerativeModels2018, dingDecipheringProteinEvolution2019} and have been used to generate novel protein sequences with predetermined functions. \citet{greenerDesignMetalloproteinsNovel2018} trained a conditional VAE, which incorporated a rough topology of the protein as an additional input, on $\sim$4,000 short monomeric structures in the PDB, as well as their homologs from UniRef, and showed that the resulting model can generate new protein sequences that correspond to a specified topology. \citet{hawkins-hookerGeneratingFunctionalProtein2021} trained a VAE on $\sim$70,000 luciferase sequences and showed that the proteins generated by the model are also often luminescent. \citet{dasAcceleratedAntimicrobialDiscovery2021} trained a VAE on peptide sequences from UniProt and, using controlled generation and screening, they were able to produce novel peptides with antimicrobial activity.

VAEs have also been used to generate backbones of proteins with predetermined topologies. \citet{eguchiIGVAEGenerativeModeling2020} trained a VAE to take as input a distance matrix and to generate 3D coordinates matching both the input distance matrix and the torsion angles of the corresponding protein structure. After training the VAE on $\sim$11,000 structures of immunoglobulins, the resulting model was able to generate novel immunoglobulin backbones matching the expected bond lengths, bond angles, and torsion angles and to learn a meaningful latent representation that could be explored to find backbones with desired shapes and characteristics.

\subsection{Normalizing flows (NFs)}

Normalizing flows (NFs) use an invertible neural network to learn a bidirectional mapping between the inputs and the latent representation (\autoref{fig:generative-models}c). The use of an invertible network makes it possible to calculate the exact probability of the training data given the model parameters and to optimize the model parameters accordingly, although it also imposes substantial constraints on the types of neural network architectures that can be used.

The most notable application of normalizing flows to protein design has been the modeling of protein dynamics \cite{noeBoltzmannGeneratorsSampling2019}. \citet{noeBoltzmannGeneratorsSampling2019} introduced Boltzmann generators: neural networks which learn a mapping between configurations of a many-body system and a latent representation. The authors showed that, after training a Boltzmann generator using a set of protein conformations and energies predicted by a molecular mechanics force-field, it is possible to generate new conformations that can be confirmed by molecular dynamics simulations, and to accurately model transitions and energy differences between known states.

\subsection{Energy-based models (EBMs)}


Energy-based models (EBMs) are a large class of models that, in lieu of learning a probability density function over the input space, are simply trained to assign low values (or ``energies'') to observed states and high values to unobserved or improbable states (\autoref{fig:generative-models}d). Training EBMs requires a strategy for efficiently sampling a representative set of improbable states, and different strategies are often employed for different applications.

EBMs have been used extensively for learning meaningful representations of protein sequences \cite{asgariContinuousDistributedRepresentation2015,yangLearnedProteinEmbeddings2018,luSelfSupervisedContrastiveLearning2020} and structures \cite{gainzaDecipheringInteractionFingerprints2020}. \citet{gainzaDecipheringInteractionFingerprints2020} introduced MaSIF, a model trained to map protein surface meshes into compact representations called ``fingerprints'' such that complementary surfaces of known binders have complementary fingerprints (i.e., have low Euclidean distance when one of the two fingerprints is negated). The resulting fingerprints can be used to perform protein-protein interaction prediction and protein docking significantly faster than traditional approaches while achieving comparable accuracy. This approach can further be extended by incorporating the feature generation step into the model architecture, allowing the model to be trained end-to-end \cite{sverrissonFastEndtoendLearning2020}.


EBMs have also been used for fixed backbone design. \citet{duEnergybasedModelsAtomicresolution2020} introduced Atom Transformer, a model trained to predict whether an amino acid rotamer matches the context defined by the identity and position of $k$ nearest atoms. The model is trained to assign low energies when contexts are paired with native rotamers and high energies when contexts are paired with non-native rotamers. The non-native rotamers are selected at random from a rotamer library, after conditioning on the backbone torsion angles and the amino acid types. The resulting model achieves comparable accuracy to Rosetta \cite{lemanMacromolecularModelingDesign2020} in recovering native rotamers, and it supports continuous rotamer representations, which would not be possible if rotamer placement was framed as a classification problem \cite{anand-achimProteinSequenceDesign2021}. However, in order to assign a rotamer to a given context, all possible rotamers have to be evaluated, which makes inference relatively slow.



\subsection{Generative adversarial networks (GANs)}

Generative adversarial networks (GANs) are a subset of EBMs where a generator network is trained to propose challenging negative examples and a discriminator network is trained to distinguish between the real and the generated examples (\autoref{fig:generative-models}e). The concomitant training of the generator network allows GANs to be efficient at generating new examples, in contrast to many other EBMs where the generation of new examples requires extensive sampling.

GANs have been used to generate \cite{anandGenerativeModelingProtein2018} and refine \cite{maddhurivenkatasubramaniyaProteinContactMap2021} distance matrices and to generate novel protein sequences with specific folds \cite{karimiNovoProteinDesign2020} and functions \cite{repeckaExpandingFunctionalProtein2021}. \citet{anandGenerativeModelingProtein2018} trained a GAN model, employing 2D convolution, pooling, and upsampling layers, to generate distance matrices corresponding to novel protein folds. Protein backbones could be reconstructed from the distance matrices either by solving a convex optimization objective \cite{anandGenerativeModelingProtein2018} or by using a model trained to map distance matrices to coordinates \cite{anandFullyDifferentiableFullatom2019}. \citet{repeckaExpandingFunctionalProtein2021} trained a GAN model, employing convolution and attention layers, on a dataset of malate dehydrogenase (MDH) sequences. Sequences generated by the resulting model were validated experimentally and possessed enzymatic activity in $\sim$24\% of cases.

\section{Model guided protein design}


In model guided protein design \cite{biswasMachineguidedDesignProteins2018}, a pretrained deep generative model, preferably conditioned on the structure \cite{ingrahamGenerativeModelsGraphBased2019a, strokachFastFlexibleProtein2020} or function \cite{madaniProGenLanguageModeling2020, alleyUnifiedRationalProtein2019a} of the target protein, is used to generate the initial pool of candidates. Discriminative oracles are then used to independently validate the generated candidates \cite{maddhurivenkatasubramaniyaProteinContactMap2021, strokachComputationalGenerationProteins2021}, to prioritize them for experimental validation \cite{biswasLowProteinEngineering2021, dasAcceleratedAntimicrobialDiscovery2021}, or to guide the generator to produce sequences or structures that are more desirable \cite{noeBoltzmannGeneratorsSampling2019, guptaFeedbackGANDNA2019, gomez-bombarelliAutomaticChemicalDesign2018, brookesConditioningAdaptiveSampling2019, nornProteinSequenceDesign2021}. Ultimately, the generative model, which can be trained on vast amounts of unlabeled data, increases the probability that the candidates correspond to valid sequences or structures, while the discriminative oracles, which can include molecular mechanics simulations or models trained on domain-specific datasets, increase the probability that the candidates have the desired functionality.





\section{Conclusions}
\label{sec:conclusions}

In this review, we described a number of protein design scenarios where deep generative models  successfully produced novel proteins \cite{ingrahamGenerativeModelsGraphBased2019a, anand-achimProteinSequenceDesign2021, rivesBiologicalStructureFunction2021} often orders of magnitude faster than traditional approaches \cite{strokachFastFlexibleProtein2020, dasAcceleratedAntimicrobialDiscovery2021}. Continued growth in the number of protein sequences and structures that are available \cite{theuniprotconsortiumUniProtUniversalProtein2021, burleyRCSBProteinData2021}, coupled with the development of protein-specific machine learning libraries \cite{jamasbGrapheinPythonLibrary2020, pavlovicImmuneMLEcosystemMachine2021}, and network architectures \cite{raoMSATransformer2021, fuchsSETransformers3D2020, kohlerEquivariantFlowsExact2020, hermosillaIntrinsicExtrinsicConvolutionPooling2021}, are likely to result in further improvements in the future.

\section{Acknowledgements}
\label{sec:acknowledgements}

AS acknowledges support from an NSERC PGS-D graduate scholarship. PMK acknowledges support from an NSERC Discovery grant (RGPIN-2017-064) and a CIHR Project grant (PJT-166008).

\section{Conflicts of interest}
\label{sec:conflicts-of-interest}

PMK is a cofounder of Resolute Bio Inc. and serves on the scientific advisory board of ProteinQure.

\appendix

\section{Model descriptions}
\label{sec:model-descriptions}

\subsection{Autoregressive models}


Autoregressive models operate on sequences of tokens and are typically trained either using the causal language modeling objective function or using the masked language modeling objective function (\autoref{fig:generative-models}a).

In causal language modelling (CLM), the goal is to predict the identity of each amino acid given the preceding amino acids in the input sequence (\autoref{eq:causal-language-modeling}). Models trained using the CLM objective are particularly well-suited for generating novel protein sequences, since this task closely resembles the objective function used to train the models.


\begin{equation}
  \label{eq:causal-language-modeling}
  \mathcal{L}_{CLM} = \mathbb{E}_{\boldsymbol{x} \sim p_{data}(\boldsymbol{x})} \left[ \log p_{\theta}(\boldsymbol{x}_0) + \sum_{i=1}^{N - 1} p_{\theta}(\boldsymbol{x}_i | \boldsymbol{x}_0, \dotsc, \boldsymbol{x}_{i - 1}) \right] \\
\end{equation}

In masked language modelling (MLM), the goal is to predict the identity for a fraction of randomly selected and masked amino acids in the input sequence (\autoref{eq:masked-language-modeling}). Models trained used the MLM objective are bidirectional and therefore are particularly well-suited for optimizing specific regions in a protein and for providing representations for each residue which capture information about both the preceding and the succeeding regions in the sequence. These models have also been used to generate entire protein sequences using sampling, beam search, or other strategies.


\begin{equation}
  \label{eq:masked-language-modeling}
  \mathcal{L}_{MLM} = \mathbb{E}_{\boldsymbol{x} \sim p_{data}(\boldsymbol{x})} \left[ \mathbb{E}_{M} \sum_i^M \log p_{\theta}(\boldsymbol{x}_i | \boldsymbol{x}_{\notin M}) \right]
\end{equation}

\subsection{Variational autoencoders (VAEs)}


Traditional autoencoders comprise an encoder network $q_{\phi}(\boldsymbol{z}|\boldsymbol{x})$, which maps an input $\boldsymbol{x}$ to a latent representation $\boldsymbol{z}$, and a decoder network $p_{\theta}(\boldsymbol{x}|\boldsymbol{z})$, which maps a latent representation $\boldsymbol{z}$ to a reconstructed input $\boldsymbol{\hat{x}}$. Variational autoencoders (VAEs) \cite{kingmaAutoEncodingVariationalBayes2013} are similar, but instead of predicting the latent variables $\boldsymbol{z}$, VAEs predict parameters of a distribution over the latent variables, or $\boldsymbol{\mu}$ and $\boldsymbol{\sigma}$ in the case where latent variables are modeled as independent Gaussians. The decoder then takes a sample from the predicted distribution, using a reparameterization trick to make the sampling process differentiable \cite{kingmaAutoEncodingVariationalBayes2013}, and maps that sample to the output $\boldsymbol{\hat{x}}$ (\autoref{fig:generative-models}b).

VAEs cannot be trained by minimizing the negative marginal probability $p_{\theta}(\boldsymbol{x})$ directly because calculating the marginal probability requires taking an integral over the latent space (\autoref{eq:vae-p}), which is intractable in most cases.

\begin{equation}
  \label{eq:vae-p}
  p_{\theta}(\boldsymbol{x}) = \int p_{\theta}(\boldsymbol{x} | \boldsymbol{z}) p(\boldsymbol{z}) dz
\end{equation}

Instead, VAEs are trained by minimizing the negative evidence lower bound (ELBO) of the data given the model parameters (\autoref{eq:vae}). The first term is the reconstruction loss between the input and the output (e.g. cross-entropy loss in the case of categorical data or mean squared error in the case of continuous data). The second term is the Kullback–Leibler distance between the predicted parameters of the latent distribution and the prior over the latent distribution (e.g. $\frac{1}{2} \sum_{j=1}^{k} \left[ \sigma_j + \mu_j^2 - 1 - \log \sigma_j \right]$ in the case where latent variables are modeled as $k$ independent Gaussians).

\begin{equation}
  \label{eq:vae}
  \mathcal{L}_{VAE}
  = \mathbb{E}_{q_{\phi}(\boldsymbol{z} | \boldsymbol{x})} \left[ \log p_{\theta}(\boldsymbol{x} | \boldsymbol{z}) \right] - D_{KL} \left( q_{\phi}(\boldsymbol{z} | \boldsymbol{x}) \mathrel{\Vert} p(\boldsymbol{z}) \right) \\
\end{equation}

\subsection{Normalizing flows (NFs)}


Normalizing flows use a bijective model $f_{\theta}^{-1}(\boldsymbol{x})$ to map inputs $\boldsymbol{x}$ to latent variables $\boldsymbol{z} \sim p(\boldsymbol{z})$ and its inverse $f_{\theta}(\boldsymbol{z})$ to map latent variables back to inputs (\autoref{fig:generative-models}c) \cite{rezendeVariationalInferenceNormalizing2015}. Using the change of variables rule, this allows the marginal probability of the data to be calculated as the marginal probability of the latent variables times the determinant of the model mapping data between the two distributions (\autoref{eq:nf-cov}).

\begin{equation}
  \label{eq:nf-cov}
  \begin{split}
    p(\boldsymbol{x})
    &= p(\boldsymbol{z}) \left| \det \frac{\partial f_{\theta}^{-1}}{\partial \boldsymbol{z}} \right| \\
    &= p(\boldsymbol{z}) \left| \det \frac{\partial f_{\theta}}{\partial \boldsymbol{z}} \right|^{-1}
  \end{split}
\end{equation}

The model $f_{\theta}$ can itself be a product of multiple submodels $f_1, \dots, f_K$ applied consecutively (\autoref{eq:nf-models}). As long as each submodel is invertible and has a tractable Jacobian determinant, the change of variables rule can be applied repeatedly to calculate the probability of each variable $\boldsymbol{z}_0, \dots, \boldsymbol{z}_K$ forming the ``flow'' (\autoref{eq:nf-flow}).

\begin{equation}
  \label{eq:nf-models}
  \boldsymbol{x} = \boldsymbol{z}_K = f_{\theta}(\boldsymbol{z}_0) = f_K \circ f_{K-1} \circ \dots \circ f_1(\boldsymbol{z}_0)
\end{equation}

\begin{equation}
  \label{eq:nf-flow}
  \log p(\boldsymbol{x}) = \log p(\boldsymbol{z}_0) - \sum_{j = 1}^K \log \left| \det \frac{\partial f_i}{\partial \boldsymbol{z}_{i - 1}} \right|
\end{equation}

The goal when training normalizing flows is to optimize the parameters of the model $f_{\theta}$ such that the negative log-likelihood of the data is minimized (\autoref{eq:nf}).

\begin{align}
  \label{eq:nf}
  \mathcal{L}_{NF} & = \mathbb{E}_{\boldsymbol{x}}\left[ -\log p(\boldsymbol{x}) \right] \\
                   & = \mathbb{E}_{\boldsymbol{x}}\left[
    -\log p(f_{\theta}^{-1}(\boldsymbol{x})) + \log \left| \det \frac{\partial f_{\theta}}{\partial \boldsymbol{z}} \right|
    \right]
\end{align}

A key limitation of normalizing flows is the need for the model $f_{\theta}$ to be invertible and for the Jacobian determinant $\left| \det \frac{\partial f_{\theta}}{\partial \boldsymbol{z}} \right|$ to be tractable. One approach to get around those limitations is to model $f_{\theta}$ as a neural network which, in each step, applies affine transformations on a fraction of the channels using information provided by the other channels \cite{dinhNICENonlinearIndependent2015, dinhDensityEstimationUsing2017a, kingmaGlowGenerativeFlow2018}. Affine transformations are invertible, and the resulting Jacobian matrix is triangular, which makes the determinant quick to calculate. Another approach is to model $f_{\theta}$ using a neural ODE \cite{chenNeuralOrdinaryDifferential2018} and to approximate the trace of the Jacobian matrix using the Hutchinson's trace estimator \cite{grathwohlFFJORDFreeformContinuous2018}. Other approaches have also been proposed \cite{behrmannInvertibleResidualNetworks2019, chenResidualFlowsInvertible2019,hoFlowImprovingFlowbased2019}, and this remains an active area of research \cite{papamakariosNormalizingFlowsProbabilistic2019, kobyzevNormalizingFlowsIntroduction2020}.


\subsection{Energy-based models (EBMs)}


The goal of energy-based models (EBMs) is to learn an energy function $E_{\theta}$ which outputs low values when the inputs correspond to probable (or ``low-energy'') states and high values when the inputs correspond to improbable (or ``high-energy'') states \cite{lecunTutorialEnergybasedLearning2006, bond-taylorDeepGenerativeModelling2021} (\autoref{fig:generative-models}d). In contrast to other approaches, the objective is not to maximize the log-likelihood of the data, which allows EBMs to sidestep the considerable challenge of approximating the partition function $Z_{\theta}$ during training (\autoref{eq:emb-boltzmann}). However, since the energies predicted by EBMs are not normalized, special care needs to be taken to make sure that the model does not assign arbitrarily low energies to all possible inputs \cite{lecunTutorialEnergybasedLearning2006}.

\begin{equation}
  \label{eq:emb-boltzmann}
  p_\theta(\boldsymbol{x})
  = \frac{\exp(-E_{\theta}(\boldsymbol{x}))}{Z_{\theta}}
\end{equation}

Contrastive divergence is a training strategy where fake examples are generated using Gibbs sampling, stochastic gradient Langevin dynamics (SGLD), or another Markov-chain Monte Carlo (MCMC) method, and the model parameters are updated such that real examples are assigned lower energies while fake examples are assigned higher energies \cite{hintonTrainingProductsExperts2002} (\autoref{eq:ebm-cd}). A major strength of contrastive divergence is that it can be applied to almost any domain while major weaknesses are computational cost of generating fake examples with every iteration and training instability due to the evolving nature of the distribution from which the fake examples are sampled.

\begin{equation}
  \label{eq:ebm-cd}
  \nabla \mathcal{L}_{CD} = \mathbb{E}_{\boldsymbol{x}_+ \sim p_{data}(\boldsymbol{x})} [\nabla E_{\theta} (\boldsymbol{x}_+)] - \mathbb{E}_{\boldsymbol{x}_- \sim p_{\theta}(\boldsymbol{x})} [\nabla E_{\theta} (\boldsymbol{x}_-)]
\end{equation}

Noise-contrastive estimation (NCE) is a training strategy where fake examples are sampled from a predefined distribution, and the model is trained to distinguish between real and fake examples akin to logistic regression \cite{gutmannNoisecontrastiveEstimationNew2010, gaoFlowContrastiveEstimation2020} (\autoref{eq:ebm-nce}). Energies $E_{\theta}(\boldsymbol{x})$ are mapped to probabilities $p_{\theta}(\boldsymbol{x})$ by introducing a learnable parameter standing for the logarithm of the partition function (\autoref{eq:emb-boltzmann}), and the key challenge of NCE is defining a tractable and representative distribution of negative examples $q(\boldsymbol{x})$. NCE can be particularly effective when the goal is to assign energies to a pairing between some categorical variable and its context, since the negative distribution can be defined by all categories that are not found in the given context. Furthermore, if the model is trained to learn vector representations for both the variable and the context, with the similarity between those vectors corresponding to the probability $p_{\theta}(\boldsymbol{x})$, then the learned embeddings are likely to carry useful information for downstream tasks.

\begin{equation}
  \label{eq:ebm-nce}
  \mathcal{L}_{NCE} = \mathbb{E}_{\boldsymbol{x}_+ \sim p_{data}(\boldsymbol{x})} \left[ \frac{p_{\theta}(\boldsymbol{x}_+)}{p_{\theta}(\boldsymbol{x}_+) + q(\boldsymbol{x}_+) }\right]
  + \mathbb{E}_{\boldsymbol{x}_- \sim q(\boldsymbol{x})} \left[ \frac{q(\boldsymbol{x}_-)}{p_{\theta}(\boldsymbol{x}_-) + q(\boldsymbol{x}_-) }\right]
\end{equation}

\subsection{Generative adversarial networks (GANs)}

A generative adversarial network (GAN) comprises a generator and a discriminator that are trained in tandem through a minimax game (\autoref{fig:generative-models}e) \cite{goodfellowGenerativeAdversarialNets2014}. Several flavors of GANs have been proposed that differ in the nonlinearities that are applied to the discriminator outputs and in the loss functions that are minimized \cite{bond-taylorDeepGenerativeModelling2021}. The simplest and most popular is the Wasserstein GAN \cite{arjovskyWassersteinGAN2017}, where the discriminator is trained to assign low values to real examples and high values to generated examples (\autoref{eq:gan-d}), while the generator is trained to produce examples that are assigned low values by the discriminator (\autoref{eq:gan-g}). GANs can be viewed as EBMs, where the discriminator corresponds to $E_{\theta}$ while the generator corresponds to the function that proposes negative examples \cite{bond-taylorDeepGenerativeModelling2021}.

\begin{equation}
  \label{eq:gan-d}
  \mathcal{L}_{D} = \mathbb{E}_{\boldsymbol{x} \sim p_{data}(\boldsymbol{x})}[D(\boldsymbol{x})] - \mathbb{E}_{\boldsymbol{z} \sim p(\boldsymbol{z})}[D(G(\boldsymbol{z}))]
\end{equation}

\begin{equation}
  \label{eq:gan-g}
  \mathcal{L}_{G} = \mathbb{E}[D(G(\boldsymbol{z}))]
\end{equation}


\bibliographystyle{elsarticle-num-names}
\bibliography{references}



\end{document}